\pdfoutput=1

\documentclass[11pt]{article}

\usepackage[]{ACL2023}
\usepackage{times}
\usepackage{latexsym}

\usepackage[T1]{fontenc}

\usepackage[utf8]{inputenc}

\usepackage{microtype}

\usepackage{inconsolata}
\usepackage{hyperref}
\usepackage{multirow}
\usepackage{graphicx}
\usepackage{tikz-dependency}
\usepackage{booktabs}
\usepackage{multirow}
\usepackage[normalem]{ulem}
\useunder{\uline}{\ul}{}
\usepackage{amsmath}

\title{Solving Label Variation in Scientific Information Extraction \\
via Multi-Task Learning}

\author{Dong Pham$^1$, Xanh Ho$^2$, Quang-Thuy Ha$^1$ \and Akiko Aizawa$^{2,3}$\\
        $^1$VNU University of Engineering and Technology, Hanoi, Vietnam \\ 
        $^2$National Institute of Informatics, Tokyo, Japan \\ 
        $^3$The University of Tokyo, Tokyo, Japan \\
         {\tt dongpham120899@gmail.com} \\
        {\tt thuyhq@vnu.edu.vn} \\
        {\tt \{xanh, aizawa\}@nii.ac.jp}
        }

\begin{document}
\maketitle

\begin{abstract}
Scientific Information Extraction (ScientificIE) is a critical task that involves the identification of scientific entities and their relationships. The complexity of this task is compounded by the necessity for domain-specific knowledge and the limited availability of annotated data. Two of the most popular datasets for ScientificIE are SemEval-2018 Task-7 and SciERC. They have overlapping samples and differ in their annotation schemes, which leads to conflicts. In this study, we first introduced a novel approach based on multi-task learning to address label variations. We then proposed a soft labeling technique that converts inconsistent labels into probabilistic distributions. The experimental results demonstrated that the proposed method can enhance the model robustness to label noise and improve the end-to-end performance in both ScientificIE tasks. The analysis revealed that label variations can be particularly effective in handling ambiguous instances. Furthermore, the richness of the information captured by label variations can potentially reduce data size requirements. The findings highlight the importance of releasing variation labels and promote future research on other tasks in other domains. Overall, this study demonstrates the effectiveness of multi-task learning and the potential of label variations to enhance the performance of ScientificIE.\footnote{Data and code are publicly available at: \url{https://github.com/dongpham120899/LabelVariation_SciIE}}
\end{abstract}

\section{Introduction}
Information extraction (IE) refers to the process of automatically identifying the entities and relations from unstructured text. Extracting the information from a scientific paper is more challenging than from open-domain data, given that scientific texts require in-depth knowledge of the subject matter for accuracy; thus, labeling is costly and the amount of labeled data is limited. \citet{bassignana-plank-2022-mean} revealed that two well-known datasets, namely, SemEval-2018 \cite{buscaldi2017semeval} and SciERC \cite{luan2018multi}, contain overlapped abstracts and directly correspondent labels; however, they differ in their annotations. In particular, there are 307 abstracts (out of 500 abstracts) that are overlapped between the two datasets. The number of annotated relations in these abstracts differs significantly and includes conflicting instances. The presence of conflicting annotations raises concerns regarding the reliability of these datasets; thus, the determination of trustworthiness is challenging, especially with limited resources.


Labeling plays a crucial role in machine learning pipelines, as it involves assigning labels to data points to train models effectively. However, human labeling is generally subject to errors and inconsistencies \cite{plank2022problem}, and label aggregation cannot capture the actual complexity of the world \cite{basile2021we}. 
Using only high-agreement instances for model training and testing can cause overfitting and data redundancy \cite{jamison2015noise}. Therefore, different annotated opinions should be retained and ``variation'' should be considered over ``disagreement'', given that disagreement annotations imply that two (or more) views involved are not all accurate \cite{plank2022problem}.

Label variation occurs in ScientificIE when different annotators assign different labels to the same entity or relationship. This variation can stem from various factors, including differences in domain knowledge or interpretations of annotation guidelines, in addition to the subjective understanding of the underlying data. In response to the challenge of label variations arising from overlapping datasets, we developed a novel approach based on multi-task learning. By jointly training the proposed model on multiple perspectives, overlapping and conflicting annotations can be effectively handled. We released soft labels (a probability distribution generated by multi-level agreements) as an auxiliary loss. Leveraging soft labels with several loss functions can reduce the penalty for errors and enhance the model robustness \cite{fornaciari-etal-2021-beyond}.

To evaluate the effectiveness of the proposed approach, we conducted experiments using overlapped data as the training set and non-overlapped data as the testing set. We compared the performances of models trained on these datasets using traditional label aggregation methods and the proposed multi-task learning approach. Additionally, we conducted a cross-dataset evaluation on the SciREX dataset \cite{jain2020scirex} and performed testing using the standard splitting of the SciERC benchmark \cite{luan2018multi}. The experimental results revealed that the proposed approach effectively mitigated the impact of label variation on model performance, thus leading to improvements in the accuracy and robustness of two of the tasks, namely, name entity recognition (NER) and relation extraction (RE). In particular, we found that label variation is particularly effective in handling ambiguous instances, and the richness of information captured by label variation can reduce data size requirements. 

Overall, the findings suggest that multi-task learning and soft labels derived from inconsistent annotations can be powerful tools for addressing label variations in ScientificIE tasks. Moreover, future research in this field should be promoted, to comprehensively consider the potential benefits of these approaches.
\section{Label Variation in ScientificIE \label{section-2}}
Inconsistencies in annotations across different datasets can result in label variation, which presents a significant challenge for accurate and reliable machine learning models. The issue of label variation is exemplified in the overlap and annotation divergence observed in the SemEval-2018 Task 7 \cite{buscaldi2017semeval} and SciERC \cite{luan2018multi} datasets, which has been discussed in previous research \cite{bassignana-plank-2022-mean}. In this section, we delve deeper into the issue of inconsistent labels and argue for the importance of releasing variation labels in ScientificIE.

\subsection{Datasets}
\paragraph{SemEval-2018 Task 7 \cite{buscaldi2017semeval}}  
This dataset\footnote{To ensure a fair comparison with SciERC, we utilized resources specific to sub-task 2 (Relation extraction and classification on clean data) from SemEval-2018 Task 7.}
comprises 500 abstracts from published research papers from the ACL Anthology. It focuses on predicting relations between two entities with six pre-defined relations (\textit{Usage, Result, Model, Part-Whole, Topic, Comparison}). The entity annotations are first automatically identified and then manually corrected by other annotators. The target is to identify maximum noun phrases, abbreviations, and their context. The relation annotation process is divided into three steps: defining, validation, and annotation. The domain experts only annotate the semantic relations that are explicit and relevant to comprehending the abstract.

\paragraph{SciERC \cite{luan2018multi}} This corpus includes annotations for scientific entities, their relations, and coreference for 500 scientific abstracts from the AI communities. They defined six types for scientific annotation entities (\textit{Method, Metric, Task, Material, Generic, OtherScientificTerm}) and seven relation types (\textit{Used-for, Evaluate-for, Feature-of, Part-of, Compare, Hyponym-of, Conjunction}). The final annotations were obtained by greedy strategy from multiple annotators. Their annotators were preferred to indicate a longer span whenever ambiguity occurs and ignore negative relations.

\begin{table}
\small
\setlength{\tabcolsep}{6pt} 
\renewcommand{\arraystretch}{1.0} 
\begin{tabular}{lll}

\toprule
                  & \textbf{SemEval-2018} & \textbf{SciERC} \\
\midrule

\multirow{5}{*}{\textit{\textbf{Label mapping}}} & Comparison   & Compare         \\
                  & Usage                       & Used-for        \\
                  & Part-whole                  & Part-of         \\
                  & Model                       & Feature-of      \\
                  & Result                      & Evaluate-for    \\
\midrule
\multicolumn{3}{l}{\textit{\textbf{Statistic on whole corpus}}}   \\
\midrule
\# Entities        &       7483       &   8089              \\
\# Relations       &       1583       &     4648            \\
\# Relations/Doc   &          3.2     &  9.3               \\
\midrule
\multicolumn{3}{l}{\textit{\textbf{Statistic on 307 overlapped abstracts}}}   \\
\midrule
\# Entities        &       4592       &   4252              \\
\# Relations       &       1087       &     2476            \\
\# Common Relations   &          1071     &  1922               \\
\bottomrule
\end{tabular}
\caption{\label{label-mapping}
Label mapping between the two datasets and statistics in both datasets.}
\end{table}

\subsection{Overlap of the Datasets}

\begin{table*}
\footnotesize
\setlength{\tabcolsep}{4.6pt} 
\renewcommand{\arraystretch}{1.3} 
\centering
\begin{tabular}{l|l}
\toprule
\multicolumn{2}{l}{\textbf{Example 1: Overlapped relation}} \\
\midrule
 &
\begin{dependency}
   \begin{deptext}[column sep=.5cm, row sep=.1ex]
        The  \& \underline{system} \&
        is based on a 
        \& \underline{multi-component architecture} \&.\\
    \end{deptext}
    \textbf{\depedge[edge unit distance=2ex]{4}{2}{\textcolor{blue}{Sem\_Usage}}}
    \textbf{\depedge[edge unit distance=1ex]{4}{2}{\textcolor{teal}{Sci\_Used-for}}}
    \wordgroup{1}{2}{2}{a0}
    \wordgroup{1}{4}{4}{a1}
\end{dependency} \\
\midrule
\multicolumn{2}{l}{\textbf{Example 2: Conflicted relation}} \\
\midrule
&
\scriptsize
\begin{dependency}
   \begin{deptext}[column sep=.5cm, row sep=.1ex]
      \underline{semantics} \&
      represented in a
      \& \underline{logical form language} \&.\\
   \end{deptext}
   \textbf{\depedge[edge unit distance=2ex]{3}{1}{\textcolor{blue}{Sem\_Model}}}
   \textbf{\depedge[edge unit distance=1ex]{3}{1}{\textcolor{teal}{Sci\_Used-fo}r}}
   \wordgroup{1}{1}{1}{a0}
   \wordgroup{1}{3}{3}{a1}
\end{dependency} \\
\midrule
\multicolumn{2}{l}{\textbf{Example 3: Conflicted entity}} \\
\midrule
SemEval-2018 & 
\scriptsize
\begin{dependency}
   \begin{deptext}[column sep=.5cm, row sep=.1ex]
      This paper introduces a
      \& \underline{system for categorizing unknown words} \&.\\
   \end{deptext}
   \wordgroup{1}{2}{2}{a0}
\end{dependency} 
\\
SciERC
&
\scriptsize
\begin{dependency}
   \begin{deptext}[column sep=.5cm, row sep=.1ex]
      This paper introduces a\& \underline{system} \&
      for
      \& \underline{categorizing unknown words} \&.\\
   \end{deptext}
   \textbf{\depedge[edge unit distance=1ex]{4}{2}{\textcolor{teal}{Sci\_Used-for}}}
   \wordgroup{1}{2}{2}{a0}
   \wordgroup{1}{4}{4}{a1}
\end{dependency} 
\\
\midrule

\midrule
\multicolumn{2}{l}{\textbf{Example 4: Different entity and relation}} \\
\midrule
SemEval-2018 & 
\scriptsize
\begin{dependency}
   \begin{deptext}[column sep=.2cm, row sep=.1ex]
      We propose a 
      \& \underline{detection method} \&.
      for orthographic variants caused by
      \& \underline{transliteration} \&.
      in a large
      \& \underline{corpus} \&.
      \\
   \end{deptext}
   \wordgroup{1}{2}{2}{a0}
   \wordgroup{1}{4}{4}{a0}
   \wordgroup{1}{6}{6}{a0}
   \textbf{\depedge[edge unit distance=1ex]{4}{6}{\textcolor{blue}{Sem\_Parth-Whole}}}
\end{dependency} 
\\
SciERC
&
\scriptsize
\begin{dependency}
   \begin{deptext}[column sep=.5cm, row sep=.1ex]
        We propose a 
      \& \underline{detection method} \&.
      for
      \& \underline{orthographic variants} \&.
      caused by
      \& \underline{transliteration} \&.
      in a large copurs .
      \\
   \end{deptext}
   \textbf{\depedge[edge unit distance=1ex]{4}{2}{\textcolor{teal}{Sci\_Used-for}}}
   \wordgroup{1}{2}{2}{a0}
   \wordgroup{1}{4}{4}{a1}
   \wordgroup{1}{6}{6}{a1}
\end{dependency} 
\\
\bottomrule

\end{tabular}
\caption{\label{label-example} The noise samples occur in label variation. For the relation part, we use ``Sem'' \textcolor{blue}{[*blue]} to denote relation in the SemEval dataset, while ``Sci'' \textcolor{teal}{[*green]} denotes the SciERC dataset.}
\end{table*}

\citet{bassignana-plank-2022-mean} identified 307 abstracts that were common to both the SemEval-2018 Task 7 and SciERC datasets. This indicates that there are 193 non-overlapped abstracts in each dataset. In addition, most of the relationships in both datasets have direct corresponding labels. To clarify the correspondence, we computed the co-occurrence score of relational labels between two pairs of entity labels, which is detailed in the Appendix \ref{appen-relation-mapping}. 


Table \ref{label-mapping} provides an overview of the label mapping and the quantities of entities and relations in each dataset. Both datasets contain an equal number of abstracts; however, there is a minor disparity in the number of entities. The significant distinction arises from the amount of annotated relations, i.e., SemEval-2018 has 3.2 relations per abstract while SciERC has 9.3 relations per abstract. We further highlighted their distinctiveness in the distribution of common relations, as shown in Figure \ref{label_distribution} in Appendix \ref{common_relations}. This inconsistency can be attributed to the different interpretations of annotation guidelines, where SemEval-2018 is focused on explicit relationships while SciERC on broader coverage


\subsection{Release the Label Variation}



The presence of label variation introduces inconsistency and ambiguity into the labeled data, which poses significant challenges for the training of accurate and reliable scientific extraction systems. Table \ref{label-example} presents four noise scenarios between the two datasets. These examples illustrate the difficulties associated with resolving significant disagreements and the limitations of dependence on the gold label. The actual world is excessively complex to be represented by an independent perspective. Nonetheless, incorporating labels from both datasets to train a single model poses a significant challenge. Section \ref{multi-learning} presents our proposed method that leverages multi-task learning to effectively address label inconsistencies arising from dataset overlaps.

\section{Multi-Task Learning to Handle with Label Variation\label{multi-learning}}

In this section, we first present a summary of the architecture of the end-to-end model for IE. We then introduce our proposed method for the multi-task learning of multi-perspectives. Finally,  we developed the soft label with multi-level agreements from inconsistent annotations to enhance the model's robustness.


\subsection{SpERT}
\citet{eberts2019span} introduced a span-based joint entity and relation extraction model referred to as SpERT, which is built upon the transformer pre-training framework. The authors highlighted the significance of localized context representation between entity pairs, which contributes to the effectiveness of their model. Furthermore, SpERT efficiently extracted a sufficient number of strong negative samples in a single BERT \cite{devlin2019bert} pass during training. Finally, SpERT outperformed previous approaches on several datasets for the joint entity and relation extraction tasks. By employing joint modeling, SpERT effectively captured dependencies between entities and their relations, thus resulting in improved performance and reduced processing time.

\subsection{Learning with Multi-Perspectives}
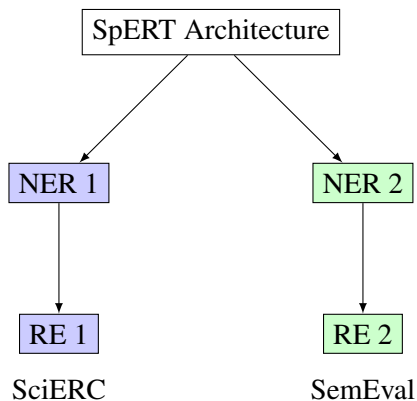
\begin{figure}[ht]
    \centering
    \begin{tikzpicture}[>=latex, node distance=2cm, every edge/.style={draw, very thick}]
    
        \node[rectangle, draw,] (input) {SpERT Architecture};
        \node[rectangle, draw, below of=input, xshift=-2cm,  fill=blue!20] (task1) {NER 1};
        \node[rectangle, draw, right of=task1, xshift=2cm, fill=green!20] (task2) {NER 2};
        \node[rectangle, draw, below of=task1,  fill=blue!20] (output1) {RE 1};
        \node[rectangle, draw, below of=task2, fill=green!20] (output2) {RE 2};
        
        \draw[->] (input) -- (task1);
        \draw[->] (input) -- (task2);
        \draw[->] (task1) -- (output1);
        \draw[->] (task2) -- (output2);

        \node[below=0.2cm of output1] {SciERC};
        \node[below=0.2cm of output2] {SemEval};
        
        
    \end{tikzpicture}
    \caption{Illustration of Multi-Task Learning based on SpERT architecture to handle label variation.}
    \label{fig:mtl-architecture}
\end{figure}

We propose an approach that utilizes two output heads in the SpERT architecture, i.e., two in NER and two in RE, where each represents a single perspective in variation annotation. This extension allows our model to address challenges such as overlapping and conflicting instances within the same input text, enabling it to learn inconsistencies in an end-to-end manner for both NER and RE tasks. To achieve this multi-perspective learning, we introduce a unified loss function that jointly optimizes entity classification and relation classification. The joint loss function is expressed as follows:
\begin{equation}
    L_i = L_i^s + L_i^r  \\
\end{equation}
\begin{equation}
    L_{multi} = L_1 + L_2
\end{equation}
where $L_i$ denotes the main loss trained with the $i$-th perspective annotation, $L^s$ denotes the span classifier loss (cross-entropy over the entity classes including none), and $L^r$ denotes the binary cross-entropy over the relation classes.\footnote{In this context, the loss function used by SciERC is denoted by $L_1$, while the loss function used by SemEval is denoted by $L_2$.} The multi-perspective loss is calculated by the sum of the single perspective loss, which encompasses both the NER and RE. To put it briefly, we used SpERT as the backbone to compute multi-perspectives, illustrated in Figure \ref{fig:mtl-architecture}.

\subsection{Soft Label from Multi-level Agreements \label{soft-label}} 
Unlike most existing models that rely on one-hot encoded gold distributions, \citet{fornaciari-etal-2021-beyond} was based on a different approach, in that probability distributions were collected over the labels provided by annotators. This allowed for a more nuanced notion of truth by comparing it with soft labels. In the overlapping examples between SemEval-2018 and SciERC, we observed both consistent and inconsistent relations, which were considered as multi-level agreements. Example 1 in Table \ref{label-example} demonstrates a high level agreement in annotations between the two datasets, whereas Example 2 illustrates conflicting relations with a low level of agreement. Several instances exhibited no similarities in entity pairs, thus resulting in different relations, as shown in Example 3. We utilized soft labels as probability distributions over the labels provided by the multi-level agreements to address these variations. 

In this study, we introduced soft labels at three levels of agreement (high, medium, and low). The soft labels were manually computed based on the degree of agreement between the two sets of data. To provide a clearer illustration, consider the label ``Sci\_Used-for'' in the first three examples in  Table \ref{label-example}. In the first example, with high agreement, the label was assigned soft labels as distributions  [0.9, 0.025, 0.025, 0.025, 0.025]. In the second example, with low agreement, the label was assigned [0.6, 0.1, 0.1, 0.1, 0.1]. Lastly, in the third example, with the medium agreement, the label was assigned probability distributions [0.8, 0.05, 0.05, 0.05, 0.05].\footnote{In this case, we only consider five types of relationship labels, with the encoding order being [Use-for, Compare, Feature-of, Part-of, Evaluate-for].}

To measure the difference between the predicted distribution Q and distribution of soft labels P, we used the Kullback-Leibler (KL) divergence \cite{kullback1951information}. The standard KL-divergence is expressed as follows:

\begin{equation}
    D_{KL}(P||Q) = \sum_{x}P(x)\log(\frac{P(x)}{Q(x)})
\end{equation}

The $D_{KL}$ describes the amount of information lost when the distribution Q is used to approximate distribution P. Moreover, the inverse KL-divergence \cite{fornaciari-etal-2021-beyond} was introduced to encourage a narrow Q distribution, which causes the model to learn a distribution that directs attention toward the classes wherein annotations exhibit potential agreement. We also attempted to compute the soft label with cross-entropy or binary cross-entropy loss, as experimentally demonstrated. 

We incorporated a soft label as an auxiliary task to mitigate overfitting. Each individual perspective was assigned its own soft label, and we included two auxiliary losses in addition to the multi-perspective loss. By considering the soft label between the two perspectives, closer alignment and facilitate learning were achieved. The final loss can be expressed as follows:
\begin{equation}
    L_{soft} = D_{KL}(P_1||Q_1) + D_{KL}(P_2||Q_2)
\end{equation}
\begin{equation}
    L_{multi\_with\_soft} = L_{multi} + L_{soft}
\end{equation}
To avoid underflow issues during computation, we applied logarithmic normalization to the soft label. Additionally, we utilized the LogSoftmax activation function for the auxiliary loss, thus ensuring that the probabilities of the individual labels did not approach zero.

\section{Experiments \label{experiments}}
To assess the effectiveness of the proposed method, we applied three main experimental scenarios to three datasets. In the first setup, we utilized the overlaps between two datasets as a training set and evaluated two of the non-overlaps as testing sets in the RE task. We then performed a cross-dataset evaluation on the SciREX dataset \cite{jain2020scirex} for the NER task. Finally, we evaluated the proposed method on the SciERC leaderboard \cite{luan2018multi} in both tasks.

For all the experiments, we leveraged the SciBERT (cased) model \cite{Beltagy2019SciBERT} as a sentence encoder, i.e., a BERT model pre-trained on a large corpus of scientific papers. We used the SpERT architecture as a baseline model to run other training sets with the same hyper-parameters reported in \citet{eberts2019span}. We utilized the spaCy toolkit \cite{spacy2} to split abstracts into sentences because both SemEval and SciERC datasets only have relations within sentences, and SpERT requires a single sentence as input. It is noted that the reported score is the average score from five runs that use different seeds.

\subsection{Overlap and Non-overlap \label{overlap and non-overlap}}
\subsubsection{Deal with Label Variation}
Two datasets with two annotation perspectives lead to inconsistencies in ScientificIE.
In this experimental setup, we utilized 307 overlapped abstracts with multiple annotation perspectives to train the model, and the two non-overlapped sets in SemEval and SciERC were used as two testing sets. However, it is important to note that SemEval-2018 did not provide entity types in their dataset. Consequently, we removed all entity types from SciERC and focused only on the RE task with five common relationships, as shown in Table \ref{label-mapping}. A performance evaluation was conducted in both sets using the micro F1-score metric, and the final score was obtained by averaging the results. 

By leveraging the corresponding labels, our primary objective was to investigate the adaptive capabilities of the two datasets through cross-evaluation (1.1 and 1.2). With the overlap, we attempted to implement other means to handle label variations in accordance with training configurations: 
\begin{itemize}
  \item \textit{concat}: We employed concatenation by including each abstract twice: once from each dataset (2.1). This technique allowed for the incorporation of sentence inputs with different annotations, thus providing a simple method to address variations in annotation \cite{sheng2008get, uma2021learning}. Additionally, to increase the amount of training data, we leveraged the non-overlapping exclusions in the respective testing sets (2.2 and 2.3).

  \item \textit{mix}: We doubled the annotation of the abstracts from two datasets. In the presence of overlapped relations, the consistent relation was retained. With the conflicting annotations, we filtered out the SemEval-2018 relation in (3.2) and the SciERC relation in (3.3).
\end{itemize}

\begin{table*}[]
\small
\setlength{\tabcolsep}{4.6pt} 
\renewcommand{\arraystretch}{1.5} 
\begin{tabular}{l|l|ccc|ccc|ccc}
\toprule
\multirow{2}{*}{No.} & Test set & \multicolumn{3}{c|}{\textbf{SemEval-2018}} & \multicolumn{3}{c|}{\textbf{SciERC}} & \multicolumn{3}{c}{\textbf{Average}} \\ \cline{2-11} 
 & Train set & \textit{Precision} & \textit{Recall} & \textit{F1-score} & \textit{Precision} & \textit{Recall} & \textit{F1-score} & \textit{Precision} & Recall & F1-score \\ \midrule
1.1 & SemEval & 21.39 & 22.93 & 22.13 & 27.30 & \hphantom{0}9.33 & 13.91 & 24.35 & 16.13 & 18.02 \\
1.2 & SciERC & \hphantom{0}6.24 & 13.84 & \hphantom{0}8.60 & 43.02 & 35.83 & 39.10 & 24.63 & 24.84 & 23.85 \\ \midrule
2.1 & Concat set & 11.73 & 23.35 & 15.62 & 41.87 & 27.99 & 33.55 & 26.80 & 25.67 & 24.59 \\
2.2 & Concat set + Sci & 10.22 & 23.55 & 14.25 & - & - & - & - & - & - \\
2.3 & Concat set + Sem & - & - & - & 39.80 & 27.36 & 32.43 & - & - & - \\ \midrule
3.1 & Mixed set & \hphantom{0}9.76 & \textbf{29.13} & 14.62 & 35.14 & 35.20 & 35.17 & 22.45 & 32.17 & 24.89 \\
3.2 & Mixed-Sci set & \hphantom{0}9.44 & 28.10 & 14.69 & 36.35 & \textbf{36.35} & 36.35 & 22.90 & \textbf{32.23} & 25.52 \\
3.3 & Mixed-Sem set & 10.10 & 28.72 & 14.95 & 34.99 & 33.68 & 33.80 & 22.54 & 31.20 & 24.28 \\ \midrule
4.1 & *MTL &  23.97 &  19.90 &  21.74 & {\textbf{44.62}} &  34.00 & 38.60 &  34.27 &  26.95 &  30.17 \\
4.2 &  *MTL with soft label &  \textbf{24.69} &  20.75 &  \textbf{22.37} &  44.33 &  35.46 &  \textbf{39.66} &  \textbf{34.51} & 28.11 &  \textbf{31.02} \\ 
\bottomrule
\end{tabular}
\caption{\label{first experiment} Micro F1-scores of the experimental training on the overlap data and testing on the non-overlap data.  1.1 and 1.2 refer to independent training data. 2.1, 2.2, and 2.3 represent the cases where we repeat abstracts from two datasets. 3.1, 3.2, and 3.3 indicate double annotation. 4.1 and 4.2 represent our proposed model.}
\end{table*}

\paragraph{Results} 
Table \ref{first experiment} reports the micro F1-scores obtained from the testing sets, including SemEval, SciERC, and their average. 
Training on independent datasets, (1.1) and (1.2) yielded satisfactory results on only one of the sets, indicating limited adaptability. The concatenation approach (2.1) to increase training data led to decreased performance due to inconsistency in the data. Further testing with non-overlapping portions of the datasets, (2.2) and (2.3) resulted in a further drop in performance.
With respect to the mixed labeling approach, we found that the mixed dataset achieved the highest recall score. Combining the two types of annotations increased the number of predictions, thus resulting in a small number of false negatives. This led to an increase in the recall score and a decrease in the precision score, which impacted the overall F1-score. Additionally, by removing conflicting relations from the mixed dataset, such as prioritizing either SciERC or SemEval annotations (3.2) or (3.3), we observed a slight improvement in performance. This emphasized the inherent limitation of traditional models in effectively capturing and incorporating multiple perspectives.
In contrast, the proposed method that utilizes multi-task learning with soft labels, achieved the optimal F1-score and precision score. It should be noted that when using multi-task learning without soft labels (4.1), the performance in individual testing sets was not superior to that when training on independent sets (Sem: 21.74 vs. 22.13 and Sci: 38.60 vs. 39.10). However, the incorporation of soft labels (4.2) improved the performance, particularly in addressing label variations and enhancing the model robustness to inconsistencies (Sem: 21.74 $\rightarrow$ 22.37 and Sci: 38.60 $\rightarrow$ 39.66). When comparing the average scores, the multi-task learning approach outperformed other methods in handling label variations within overlapped datasets.

\subsubsection{Impact of Data Quantity}

The success of training a deep learning model relies significantly on the availability of sufficient training data. In the context of ScientificIE, the limited availability of data can be attributed to the requirement of expert labeling. Obtaining a sizeable amount of such variations in label data is increasingly challenging (1400 sentences in 307 overlapped abstracts). In a previous study \cite{zhang2021learning}, new annotation distribution schemes were investigated with respect to the learning of multiple labels per example for a small subset of training examples, which can lead to novel architectures. Moreover, \citet{plank2022problem} revealed the potential of label variations to reduce data size. Thus, we conducted a comparison within the SciERC testing set between the gold and variation labels when decreasing data quantity. 

\begin{figure}[htbp]
\centering
\hspace*{-0.5cm}
\includegraphics[width=8cm]{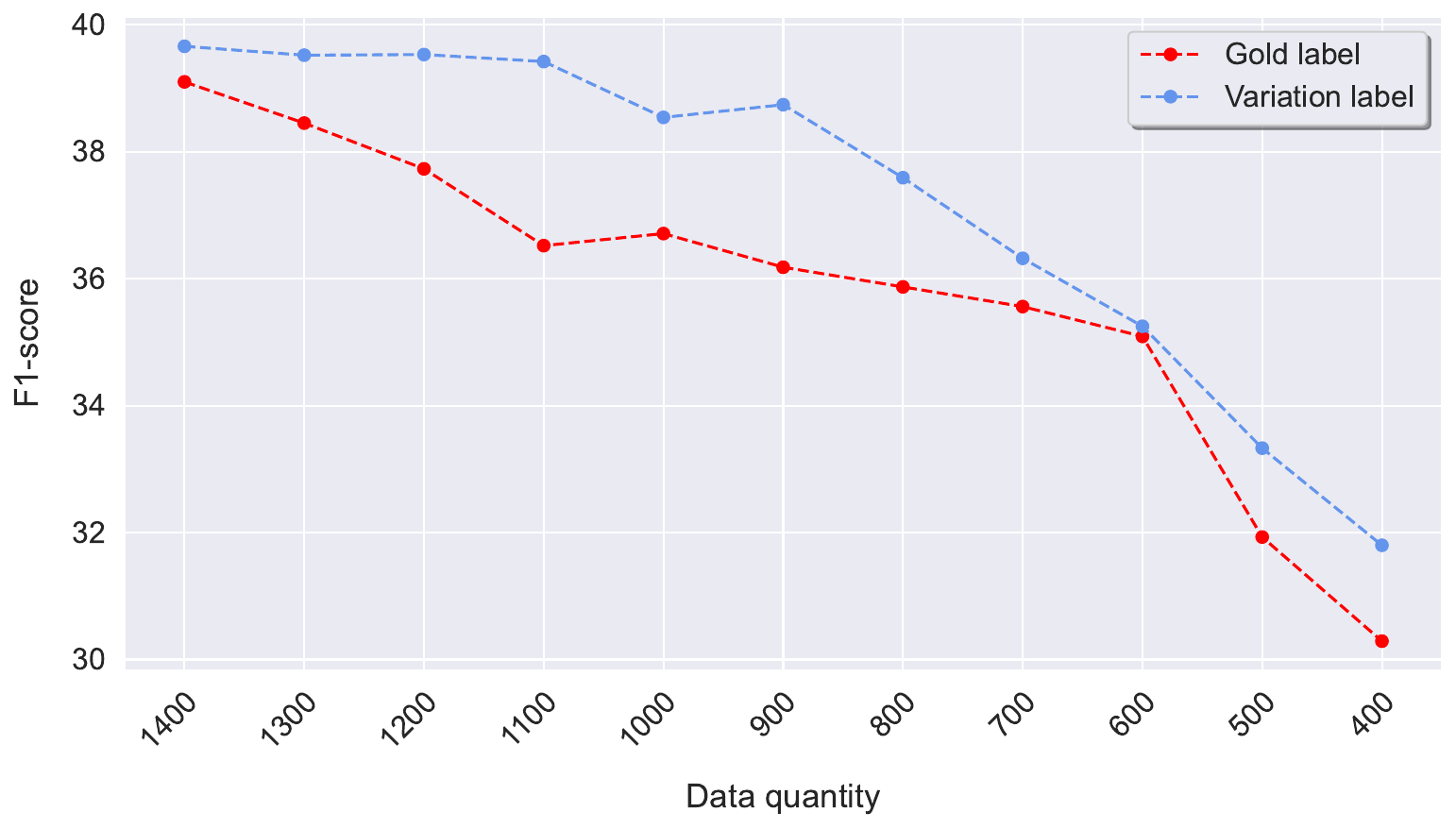}
\caption{The impact of data quantity on the performance (SciERC testing). The gold label was trained on SciERC annotations with the SpERT baseline model, and the variation label was trained using the proposed model.}
\label{drop_data_distribution}
\end{figure} 

\paragraph{Results}
Figure \ref{drop_data_distribution} illustrates the downward trend in performance in accordance with a decrease in the quantity of data in intervals of 100 training samples. The performance of the gold labels (shown in red) exhibited a rapid decrease when the data amount was reduced from 1400 to 1100. In contrast, the variation labels trained using the proposed method (shown in blue) exhibited minimal changes within the same range. Both methods exhibited a similar decline in performance when the dataset size was excessively small (from 400–1000 samples). This observation demonstrated that the richness of information captured by label variations remained stable, even with smaller datasets. Achieving a tradeoff between the data quantity and label diversity is a critical consideration for maximizing the effectiveness of machine learning models in various tasks.

\subsubsection{Impact of Soft Label}

\begin{table}[]
\small
\setlength{\tabcolsep}{6pt} 
\renewcommand{\arraystretch}{1.5} 
\begin{tabular}{l|ccc}
\toprule
\multicolumn{1}{c|}{\textbf{Model}} & \multicolumn{3}{c}{\textbf{RE}} \\ \cline{2-4} 
\multicolumn{1}{c|}{} & \multicolumn{1}{c|}{\textit{Precision}} & \multicolumn{1}{c|}{\textit{Recall}} & \textit{F1-score} \\ \midrule
MTL & 44.62 & 34.00 & 38.60 \\ \midrule
MTL + BCE & 41.94 & 34.74 & 38.01 \\ \midrule
MTL + CrossEntropy & 42.65 & 34.86 & 38.36 \\ \midrule
MTL + KL-inverse & \textbf{45.76} & 34.97 & 39.65 \\ \midrule
MTL + KL-standard & 44.33 & \textbf{35.46} & \textbf{39.66} \\ 
\bottomrule
\end{tabular}
\caption{\label{loss_function experiment} The performance of multi-task learning with (MTL row) and without soft labels (the rest). We compared the effects of different loss functions for soft labels, including the following: BCE, cross-entropy, KL-inverse \cite{fornaciari-etal-2021-beyond}, and KL-standard. All the scores in this table are obtained by evaluating the model on the SciERC testing set.}
\end{table}

We presented the soft labeling developed from multi-level agreements, as shown in Section \ref{soft-label}, and used other loss functions to capture the soft loss. \citet{fornaciari-etal-2021-beyond} proposed an inverse version of the KL-divergence, which was unsuccessful when we applied the logarithmic norm; thus, we computed the inverse version without normalization. Moreover, we attempted a standard cross-entropy with a softmax activation function at the final output head and a standard Binary-Cross-Entropy (BCE) with a sigmoid layer at the output head. 

\paragraph{Results}
Table \ref{loss_function experiment} reports the micro-F1 scores on the SciERC testing set of the MLT model with and without soft label. 
Using BCE and cross-entropy to measure the difference between the prediction distribution and the distribution of soft labels reduced the performance when compared with the case wherein soft labels were not used (only MTL). 
In this study, the inverse version of KL-divergence was not superior to the standard version with logarithmic normalization. There were few differences between the two versions. Overall, KL led to consistent performance improvements in soft labels.

\subsection{Cross-Dataset Evaluation on SciREX \label{cross-dataset experiment}}

SciREX \cite{jain2020scirex} is a document-level IE dataset for scientific articles, which covers tasks such as entity identification and N-ary relation extraction. In particular, it combines automatic and human annotations, thus leveraging existing scientific knowledge resources.

In the Table \ref{label-mapping}, there are still differences in entity annotations between the two datasets. SciERC annotators indicate the long span, including prepositions. While entity annotations in SemEval-2018 are indicated maximal noun phrases, abbreviations, etc., they often are shorter (Example 3 or 4 in Table \ref{label-example}). To evaluate the cross-dataset, we selected 368 abstracts from full-text papers in the SciREX dataset and investigated the NER task on four entity types (Method, Task, Metric, and Material) released by the SciREX dataset. We only used the abstract section, given that the SemEval or SciERC dataset only uses the abstract. We trained the SpERT model on gold labels with SciERC annotations. With variations, we retained the entity annotations of SciERC (six entity types) and released a new entity type denoted as ``OtherScientificTerm\_2''{\footnote{SemEval-2018 didn't release entity types in their dataset.} for SemEval-2018 annotations. Entity prediction results were obtatined at the head of the SciERC prediction, and the final scores were calculated by micro-averaging four entity types.
%

\begin{table}[]
\small
\begin{center}
\setlength{\tabcolsep}{4pt} 
\renewcommand{\arraystretch}{1.3} 
\begin{tabular}{lc|c}
\toprule
\multicolumn{1}{c}{\multirow{2}{*}{\textbf{Model}}} & \multirow{2}{*}{\textbf{Train set}} & \textbf{NER} \\
\multicolumn{1}{c}{} &  & \textit{F1-score} \\ \midrule
SpERT & SciERC's gold label & 43.31 \\ \midrule
SpERT\_MTL & Variation label & 44.37 \\ \midrule
SpERT\_MTL + soft label & Variation label & \textbf{44.64} \\ 
\bottomrule
\end{tabular}
\caption{\label{SciREX experiment} Micro F1-score of cross-dataset evaluation in the NER task.}
\end{center}
\end{table}

\paragraph{Results}
We compared the performances between the gold labels and variation labels of SciERC annotations, and the results revealed in Table \ref{SciREX experiment}. The two models trained on label variations outperformed the model trained on the gold labels (>1\%). This highlights the effectiveness of combining two types of annotations from the overlap, as it leads to improved performance compared with using a single annotation. Additionally, we observed the impact of using soft labels in this experiment, thus further emphasizing their effectiveness in addressing label inconsistencies.

\begin{table}[ht]
\small
\centering
\setlength{\tabcolsep}{4.2pt} 
\renewcommand{\arraystretch}{1.3} 
\begin{tabular}{llcc}
\toprule
\multicolumn{1}{c}{\multirow{2}{*}{\textbf{Model}}} & \multicolumn{1}{l|}{\multirow{2}{*}{\textbf{Label}}} & \multicolumn{1}{c|}{\textbf{NER}} & \textbf{RE} \\ \cline{3-4} 
\multicolumn{1}{c}{} & \multicolumn{1}{l|}{} & \multicolumn{1}{c|}{\textit{F1-score}} & \textit{F1-score} \\ \midrule
\textit{\textbf{Pipeline model}} &  &  &  \\ \midrule
PL.Marker & \multicolumn{1}{l|}{gold} & \multicolumn{1}{c|}{69.90} & \textbf{53.20} \\ \midrule
\textit{\textbf{End-to-End model}} &  &  &  \\ \midrule
{\ul SpERT\_MTL + soft label} & \multicolumn{1}{l|}{variation} & \multicolumn{1}{c|}{{\ul \textbf{70.83}}} & {\ul 51.31} \\
SpERT.PL & \multicolumn{1}{l|}{gold} & \multicolumn{1}{c|}{70.53} & 51.25 \\
{\ul SpERT\_MTL} & \multicolumn{1}{l|}{variation} & \multicolumn{1}{c|}{{\ul 70.61}} & {\ul 51.02} \\
SpERT & \multicolumn{1}{l|}{gold} & \multicolumn{1}{c|}{70.30} & 50.84 \\
\bottomrule
\end{tabular}
\caption{\label{SciERC experiment} The comparison of existing methods on the leaderboard of SciERC, underlined is our method.}
\end{table}

\subsection{Standard Splitting SciERC}

The SciERC benchmark \cite{luan2018multi} consists of two sets: a training set and a testing set. Among 307 overlapped abstracts, 252 abstracts are included in the training set (400 abstracts) and 55 in the testing set (100 abstracts). Due to this overlap, we were unable to obtain variation labels for the entire training dataset, which can be considered as a disadvantage. Using the proposed method, we trained both the overlapping and non-overlapping samples using two tasks and two types of annotations. Among the 148 non-overlapping abstracts, we retained those with medium-level agreement in the soft labels. The experimental setup was identical to the one described in section \ref{cross-dataset experiment}. The micro F1-scores for both tasks were calculated based on the gold label annotations from SciERC.{\footnote{In this experiment, we considered the whole entity and relation types of both datasets.}}

\paragraph{Results}
The performance of the proposed method on the SciERC leaderboard is presented in Table \ref{SciERC experiment}. The proposed approach surpassed the state-of-the-art models in entity recognition, thus achieving an improvement of 0.6–0.8\% when compared with the SpERT \cite{eberts2019span} model and its variant, i.e., SpERT.PL \cite{santosh2021joint}. The incorporation of diverse entity annotations, even with minor conflicts, is beneficial for enhancing the accuracy of the NER task. In the RE task, the proposed method achieved the highest F1-score among existing end-to-end models. However, the improvements were limited due to significant conflicts in the relation annotations between the two datasets and the testing set as the gold label of SciERC. Furthermore, the proposed approach did not outperform a pipeline model \cite{ye2022plmarker} in the RE task. Overall, the proposed method, which utilizes multi-task learning with soft labels based on the SpERT architecture, enhanced the performance of the baseline model in both tasks.

\section{Error Analysis}

\begin{table*}[t]
\footnotesize
\small
\renewcommand{\arraystretch}{1} 
\scriptsize
\centering
\begin{tabular}{l|l}
\toprule
\multicolumn{2}{l}{\textbf{(a) Common Error} \label{error-sample-a}} \\
\midrule
\centering
\textit{\textbf{Example 1}} & 
\scriptsize
\begin{dependency}
   \begin{deptext}[column sep=.2cm, row sep=.1ex]
        \& \textcolor{red}{Sci\_Task} \&  \&  \textcolor{red}{Sci\_Task} \\
      ... whether they believed the sample output to be an \&
       \underline{expert human translation} \&
      on a \&
       \underline{machine translation} \&.\\
        \& \textcolor{teal}{Sci\_OST} \&  \&  \textcolor{teal}{Sci\_OST} \\
   \end{deptext}
   \textbf{\depedge[edge unit distance=1ex]{2}{4}{\textcolor{red}{Conjunction}}} 
   \wordgroup{1}{2}{2}{a0}
   \wordgroup{1}{4}{4}{a1}
   \wordgroup{3}{2}{2}{a2}
   \wordgroup{3}{4}{4}{a3}
   \groupedge[edge below]{a2}{a3}{\textcolor{teal}{Compare}}{1ex}
\end{dependency} 
\\
\midrule
\centering
\textit{\textbf{Example 2}} & 
\scriptsize
\begin{dependency}
   \begin{deptext}[column sep=.2cm, row sep=.1ex]
        \& \textcolor{red}{Sci\_Task} \&  \\
      We present results on\&
       \underline{addressee identification} \&
      in four-participants face-to-face meetings ..
       \\
        \& \underline{addressee identification in four-participants face-to-face meetings} \& ...  \\
        \& \textcolor{teal}{Sci\_Task} \&  \\
   \end{deptext}
   \wordgroup{1}{2}{2}{a0}
   \wordgroup{4}{2}{2}{a2}
\end{dependency}
\\
\midrule

\midrule
\multicolumn{2}{l}{\textbf{(b) Redundant Error}} \\
\midrule
\textit{\textbf{Example 3}}  & 
\scriptsize
\begin{dependency}
   \begin{deptext}[column sep=.2cm, row sep=.1ex]
       \&\textcolor{red}{Sci\_Material}\& \&\textcolor{red}{Sci\_Metric} \&  \& \textcolor{red}{Sci\_Metric}\& \& \textcolor{red}{Sci\_Metric} \\
      Our preliminary experiments on building a
      \& \underline{paraphrase corpus} \&.
      ...
      \& \underline{cost-efficiency} \&
      ,
      \& \underline{exhaustiveness} \&
      , and
      \& \underline{reliability} \&
      .
      \\
      \&\textcolor{teal}{Sci\_Material}\& \& \&  \& \& \&  \\
      \&\textcolor{blue}{Sem\_OST}\& \&\textcolor{blue}{Sem\_OST} \&  \& \textcolor{blue}{Sem\_OST}\& \& \textcolor{blue}{Sem\_OST} \\
   \end{deptext}
   \wordgroup{1}{2}{2}{a0}
   \wordgroup{1}{4}{4}{a0}
   \wordgroup{1}{6}{6}{a0}
   \wordgroup{1}{8}{8}{a0}
   \wordgroup{3}{2}{2}{a0}
   \wordgroup{4}{2}{2}{a0}
   \wordgroup{4}{4}{4}{a0}
   \wordgroup{4}{6}{6}{a0}
   \wordgroup{4}{8}{8}{a0}
   
   \textbf{\depedge[edge unit distance=1ex]{4}{6}{\textcolor{red}{Conjunction}}}
   \textbf{\depedge[edge unit distance=1ex]{6}{8}{\textcolor{red}{Conjunction}}}
\end{dependency} 
\\
\midrule

\multicolumn{2}{l}{\textbf{(c) Confusing Label}} \\
\midrule
\textit{\textbf{Example 4}}  & 
\scriptsize
\begin{dependency}
   \begin{deptext}[column sep=.2cm, row sep=.1ex]
       \&\textcolor{red}{Sci\_OST}\& \&\textcolor{red}{Sci\_Metric} \&  \\
     Both 
      \& \underline{learners} \&.
      perform well, yielding similar
      \& \underline{success rates} \&
      of approx 90 \% .
      \\
      \&\ \& \&\textcolor{teal}{Sci\_Metric} \&  \\
      \& \& \&\textcolor{blue}{Sem\_OST} \&  \\
   \end{deptext}
   \wordgroup{1}{2}{2}{a0}
   \wordgroup{1}{4}{4}{a0}
   \wordgroup{3}{4}{4}{a0}
   \wordgroup{4}{4}{4}{a0}
   
   \textbf{\depedge[edge unit distance=1ex]{2}{4}{\textcolor{red}{Evaluate-for}}}
\end{dependency} 
\\
\midrule

\textit{\textbf{Example 5}}  & 
\scriptsize
\begin{dependency}
   \begin{deptext}[column sep=.2cm, row sep=.1ex]
       \&\textcolor{red}{Sci\_Generic}\& \&\textcolor{red}{Sci\_OST} \& \& \textcolor{red}{Sci\_Material} \& \&\textcolor{red}{Sci\_OST} \& \&\textcolor{red}{Sci\_Material} \\
        We consider two groups of 
      \& \underline{indicators} \&.
      :
      \& \underline{post level} \&
      (determined ...  
      \& \underline{blog posts} \&
      only) and
      \& \underline{blog level} \&
      (determined ...  
      \& \underline{blogs} \&
      ).  
      
      \\
      \&\textcolor{teal}{Sci\_Generic}\& \& \& \& \textcolor{teal}{Sci\_Material} \& \& \& \&\textcolor{teal}{Sci\_Material} \\
      \&\textcolor{blue}{Sem\_OST}\& \& \& \& \textcolor{blue}{Sem\_OST} \& \& \& \&\textcolor{blue}{Sem\_OST} \\
   \end{deptext}
   \wordgroup{1}{2}{2}{a0}
   \wordgroup{1}{4}{4}{a0}
   \wordgroup{1}{6}{6}{a0}
   \wordgroup{1}{8}{8}{a0}
   \wordgroup{1}{10}{10}{a0}
   \wordgroup{3}{2}{2}{a0}
   \wordgroup{3}{6}{6}{a0}
   \wordgroup{3}{10}{10}{a0}
   \wordgroup{4}{2}{2}{a0}
   \wordgroup{4}{6}{6}{a0}
   \wordgroup{4}{10}{10}{a0}
   
   \textbf{\depedge[edge unit distance=1ex]{2}{4}{\textcolor{red}{Hyponym-of}}}
   \textbf{\depedge[edge unit distance=1ex]{2}{8}{\textcolor{red}{Hyponym-of}}}
   \textbf{\depedge[edge unit distance=1ex]{4}{6}{\textcolor{red}{Used-for}}}
   \textbf{\depedge[edge unit distance=1ex]{8}{10}{\textcolor{red}{Used-for}}}
\end{dependency} 
\\
\bottomrule

\end{tabular}
\caption{\label{error-analysis} The error samples are from the predictions from the proposed method: (a) the common sources of error, (b) predictions are redundant the relations, and (c) correct predictions are missing in the gold label. \textcolor{red}{[*red]} is predictions, \textcolor{teal}{[*green]} is the gold label of SciERC, \textcolor{blue}{[*blue]} is the gold label of SemEval-2018, OST is ``OtherScientificTerm''.}
\end{table*}



In this section, we conducted a detailed error analysis to gain insights into the limitations and potential areas for improvement of the proposed model. Table \ref{error-analysis} contains three error cases in the testing set. It should be noted that ScientificIE is a challenging task, and the proposed model exhibited common errors, as outlined in Example 1 (wrong entity type and relation type) or in Example 2 (incorrect spans). In certain instances, the proposed model successfully identified entities indicated in SemEval-2018 and not in SciERC. However, it tended to overpredict the relationship between these entity pairs (which is not an inaccurate relationship), as shown in Example 3. Besides, we observed that the correct entity predictions were missing in the gold labels (``learners'' in Example 4 , or ``post level'' and ``blog level'' in Example 5). The relations between entity pairs were then accurately identified. The proposed model demonstrates the capacity to cover all entities and their relationships, including the most challenging and ambiguous cases. In contrast, the gold label annotations may not always capture these complex instances accurately.

\section{Related Work}

ScientificIE systems can be developed using two main approaches: separate models, where entity extraction and relation extraction are treated as independent tasks with separate models trained for each \cite{xiao-etal-2020-denoising, zhong-chen-2021-frustratingly, ye2022plmarker}, and joint models (end-to-end models) that tackle both tasks simultaneously \cite{eberts2019span, luan-etal-2019-general, santosh2021joint}.

In recent years, various studies have prompted the community to explore innovative approaches to data labeling based on label variation \cite{passonneau2010word, plank2014learning, basile2021we, gordon2021disagreement, leonardelli2021agreeing, prabhakaran-etal-2021-releasing, uma-etal-2021-semeval, bassignana2022crossre, plank2022problem}. In this context, the proposed model leveraged variation labels in ScientificIE, thus demonstrating improved robustness with respect to label noise, in addition to higher performances in both tasks. To handle inconsistent labels and mitigate label noise, soft labeling techniques were introduced, such as the probabilistic soft labeling framework proposed by \citet{fornaciari-etal-2021-beyond}.

\section{Conclusion}

Label variation in ScientificIE introduces inconsistencies and ambiguities to labeled data, thus posing significant challenges for the training of accurate and reliable systems in this field. To overcome these challenges, we propose a multi-task learning approach that effectively handles label variations. By incorporating soft labels generated through multi-level agreements, we observed improvements in the performances demonstrated in entity and relation extraction tasks. The results indicate that label variations capture rich information and exhibit the potential to reduce data size requirements. Moreover, label variations are effective in handling ambiguous instances. The findings emphasize the significance of considering label variations in ScientificIE, and further promote its investigation in other domains and tasks.

\section*{Limitations}

This study acknowledges several limitations that should be considered. First, the findings are based on a small dataset comprised of published research papers, which may limit the generalizability of the results to a larger population or different contexts. Second, the generation of accurate and reliable soft labels remains a challenge, as the manual setting of probability distributions introduces subjectivity. Additionally, the evaluation of the experiments solely relying on the F1-score using gold labels may be impacted by errors and inconsistencies within the gold label annotations, as revealed in the error analysis. Finally, it is essential to note that this work is primarily focused on the scientific domain, and the prevalence of conflict cases may differ in other domains, thus limiting the direct transferability of the findings.

Future research should address these limitations by incorporating larger and more diverse datasets, improving the methodology for generating soft labels, considering multiple evaluation metrics, and investigating the performances of large language models in ScientificIE tasks.

\section*{Acknowledgments}
This work was (partly) supported by JSPS KAKENHI Grant Number 22K19818, and JST, AIP Trilateral AI Research, Grant Number JPMJCR20G9, Japan.

\bibliography{anthology,custom}
\bibliographystyle{acl_natbib}

\newpage
\appendix

\section{\label{appen-relation-mapping}Label Mapping}

The datasets SemEval-2018 Task 7 and SciERC contain directly corresponding labels, as detailed in Section \ref{section-2}. To establish this correspondence, we compared the co-occurrence distribution of related relation labels between entity pairs in the 307 overlapping abstracts. To ensure consistency, we transferred entity labels with the same boundaries from SciERC, as SemEval-2018 Task 7 did not release entity types. With entity types only in SemEval, we retained type ``OtherScientifiTerm\_2''. The co-occurrence score was computed using the following formula:

\begin{equation}
    O(i,j,k) = \frac{A(e_i^1, e_j^2)^{r_k}}{N_i^1 + N_j^2}
\end{equation}

where 
$ A(e_i^1, e_j^2)^{r_k} $ denotes the number of occurrence relations $r_k$ between entity pairs $e_i^1$ and $e_j^2$.
$N_i^1, N_j^2$ represent the number of occurrences of entity label $i, j$ was entity $1, 2$. The specific comparisons and descriptions can be found in Figures \ref{occurrence-Compare}, \ref{occurrence-Use-for}, \ref{occurrence-Evaluate-for}, \ref{occurrence-Feature-of}, and \ref{occurrence-Part-of}. Corresponding relations exhibited similar co-occurrence distributions, with certain relations such as ``Compare'' and ``Comparison'' appearing most frequently between entity pairs such as ``Method'' and ``Method'' or ``Task'' and ``Task''.

\section{\label{common_relations} Common Relations}

\begin{figure}[htbp]
\centering
\includegraphics[width=8cm]{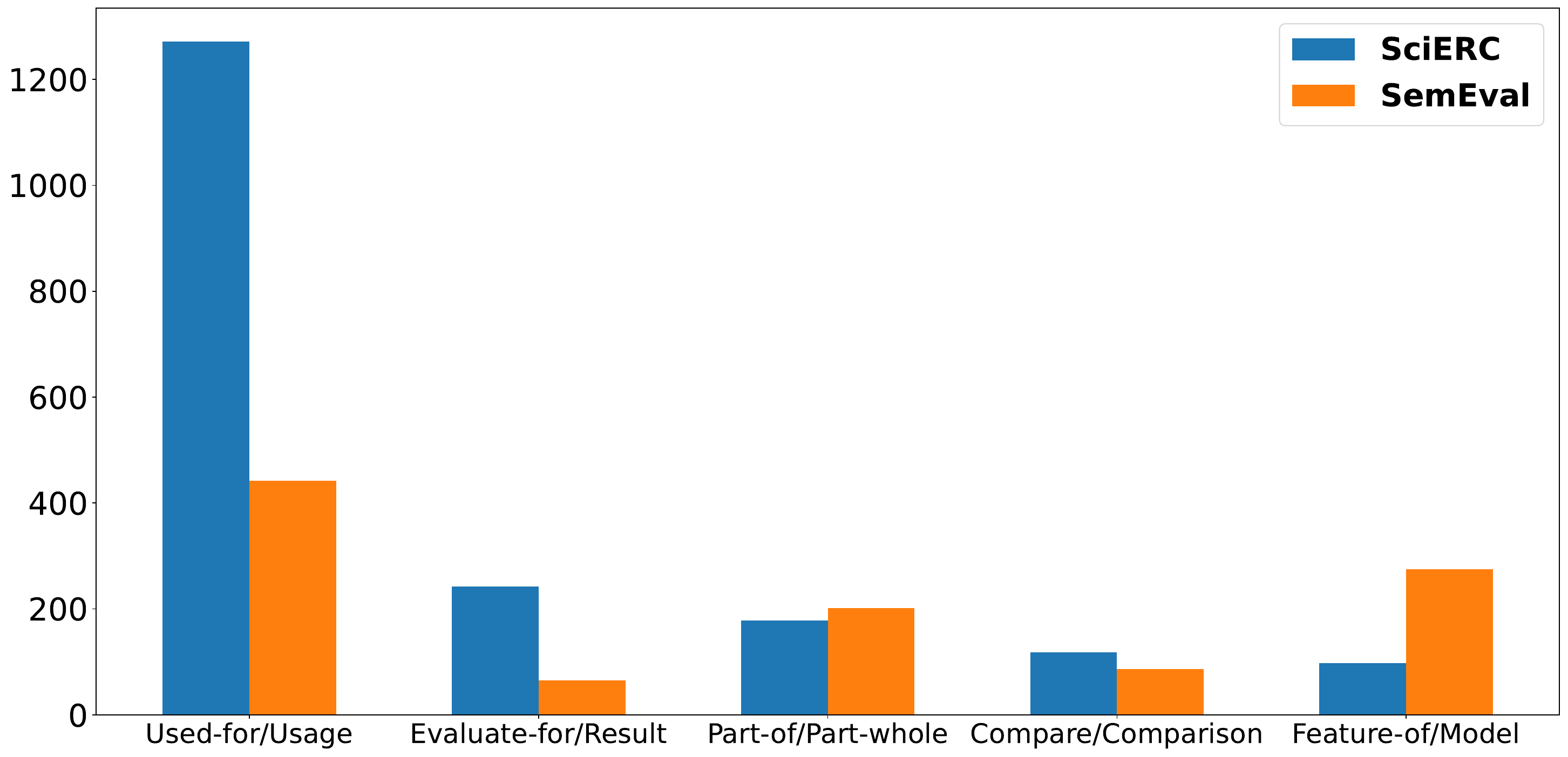}
\caption{The distribution of common relations in the overlapped abstracts of the two datasets.}
\label{label_distribution}
\end{figure}

We observed that most relations in both datasets are labeled as \textit{``Used-for/Usage''}. However, there is a notable disparity between the two datasets regarding label distribution. Specifically, in SemEval, labels such as \textit{``Model''} and \textit{``Part-whole''} have a significantly larger number of occurrences when compared with their counterparts in SciERC.

\begin{figure*}[htbp]
\centering
\includegraphics[width=16cm]{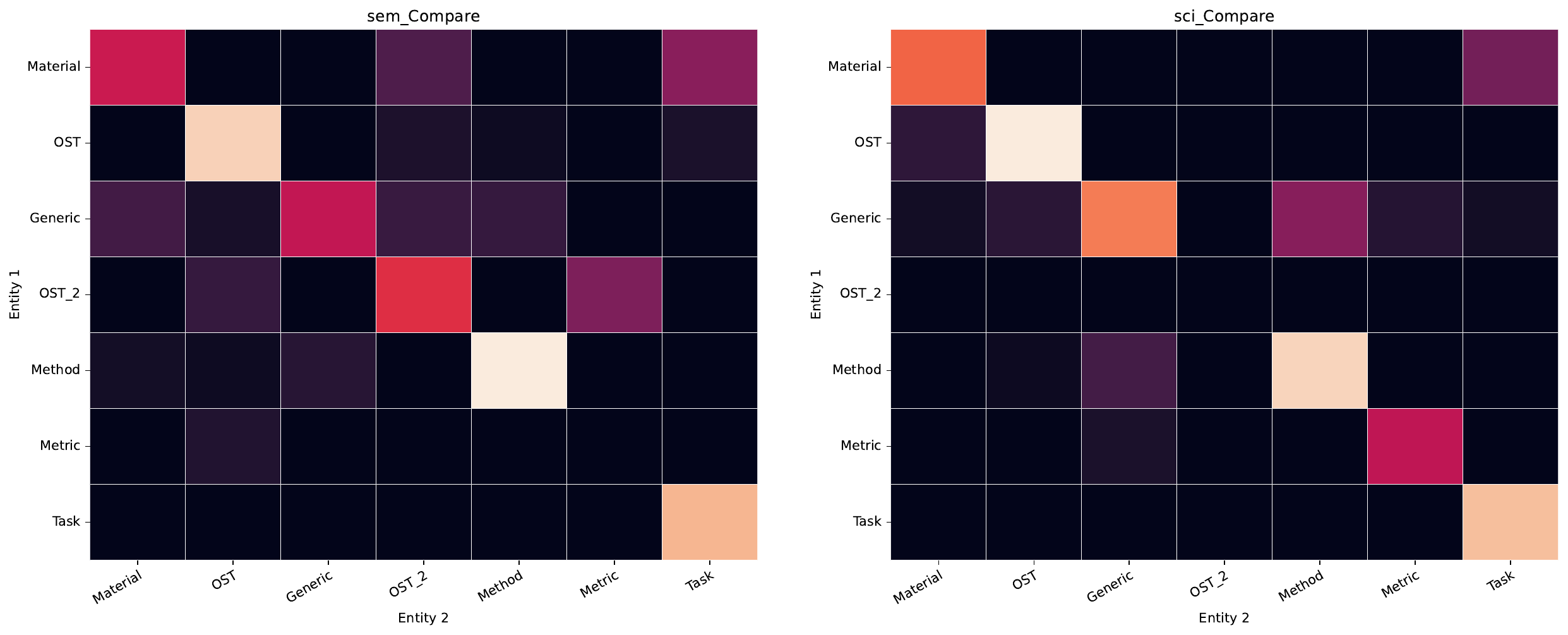}
\caption{The co-occurrence distribution of ``Compare/Comparision''  in two overlapped corpora.}
\label{occurrence-Compare}
\end{figure*} 

\begin{figure*}[htbp]
\centering
\includegraphics[width=16cm]{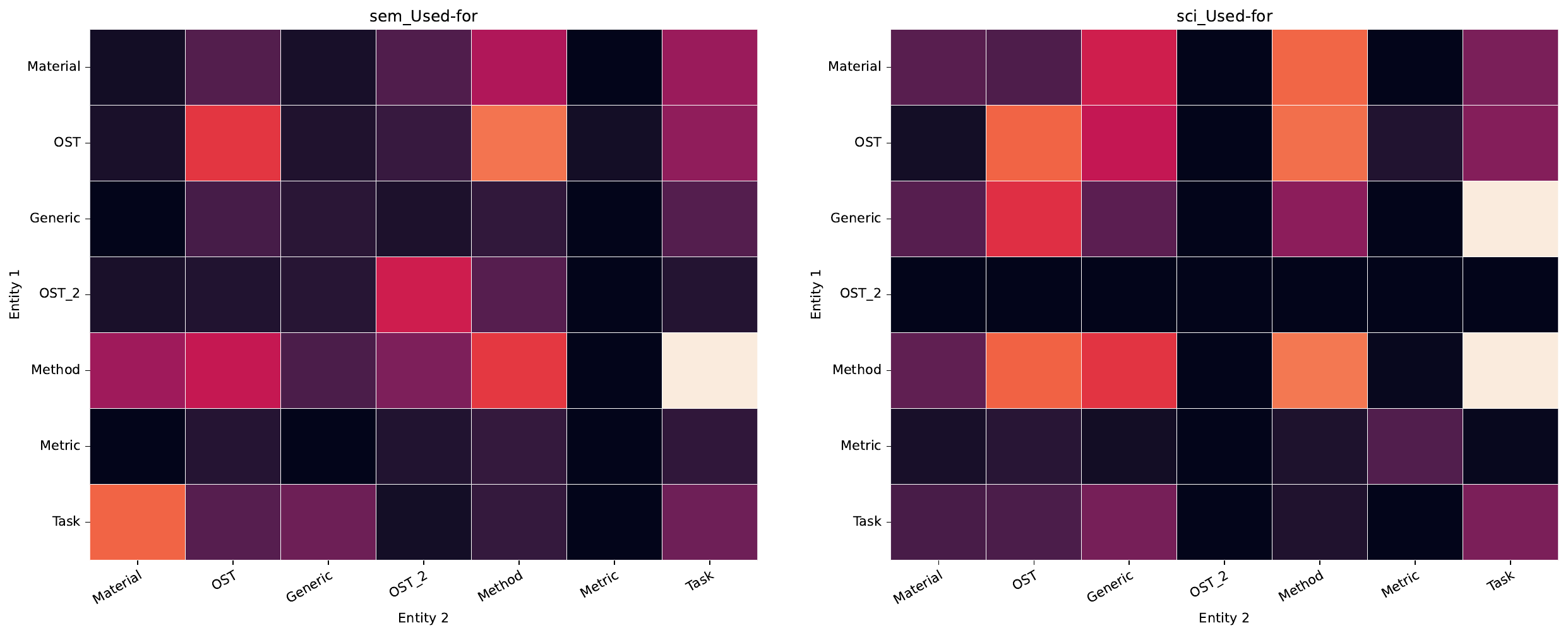}
\caption{The co-occurrence distribution of ``Used-for/Usage''  in two overlapped corpora.}
\label{occurrence-Use-for}
\end{figure*} 

\begin{figure*}[htbp]
\centering
\includegraphics[width=16cm]{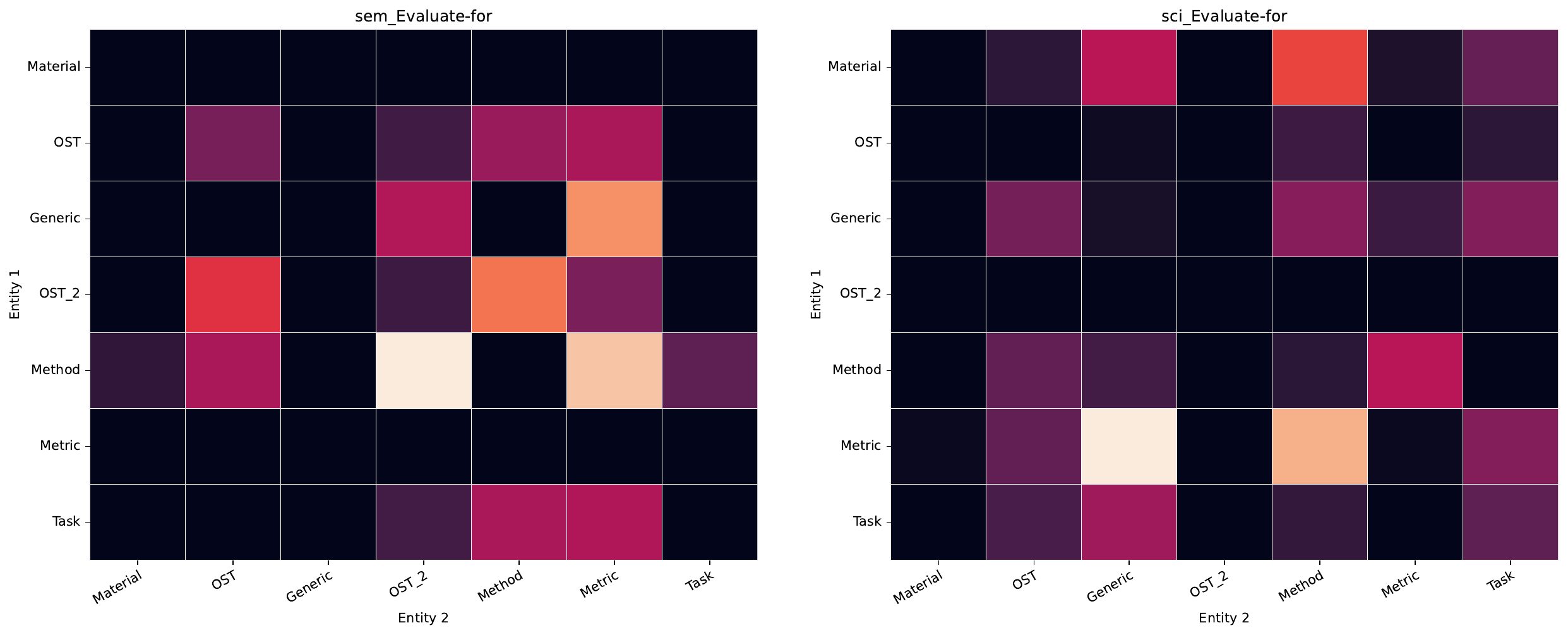}
\caption{The co-occurrence distribution of ``Evaluate-for/Result''  in two overlapped corpora.}
\label{occurrence-Evaluate-for}
\end{figure*} 

\begin{figure*}[htbp]
\centering
\includegraphics[width=16cm]{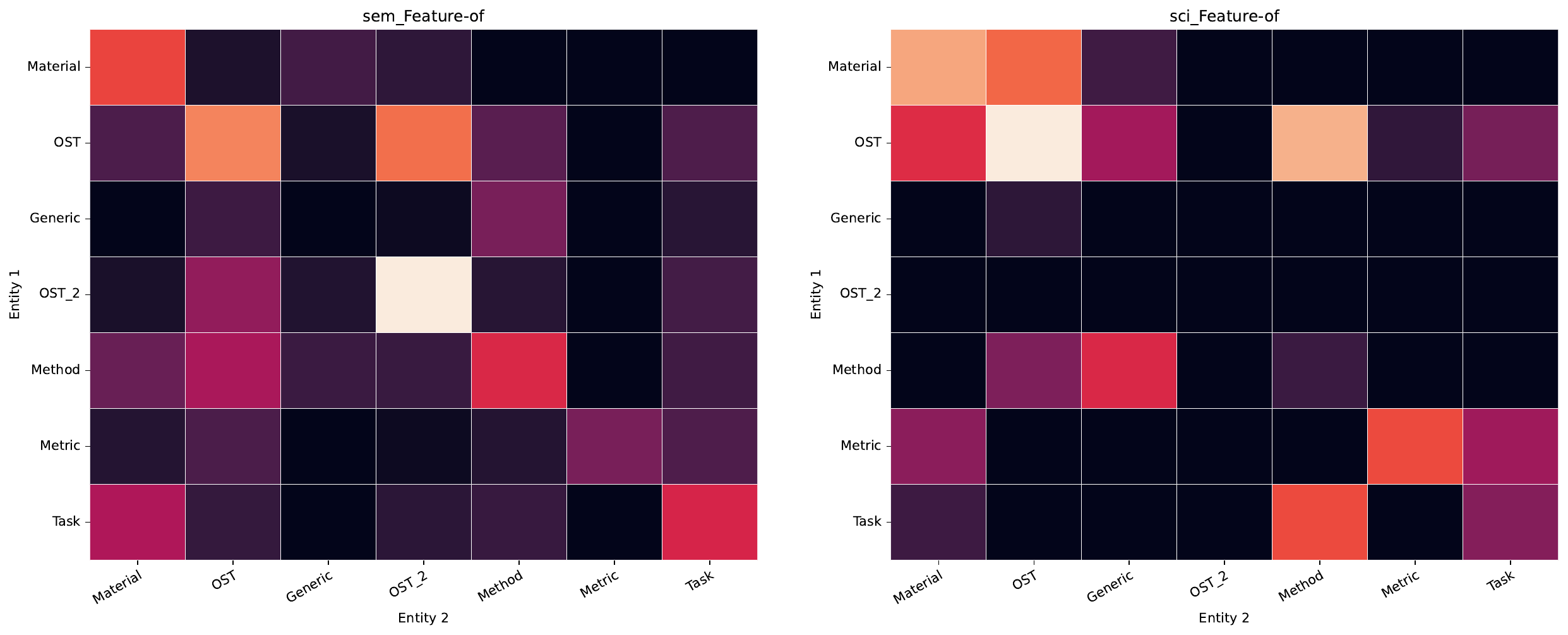}
\caption{The co-occurrence distribution of ``Feature-of/Model''  in two overlapped corpora.}
\label{occurrence-Feature-of}
\end{figure*} 

\begin{figure*}[htbp]
\centering
\includegraphics[width=16cm]{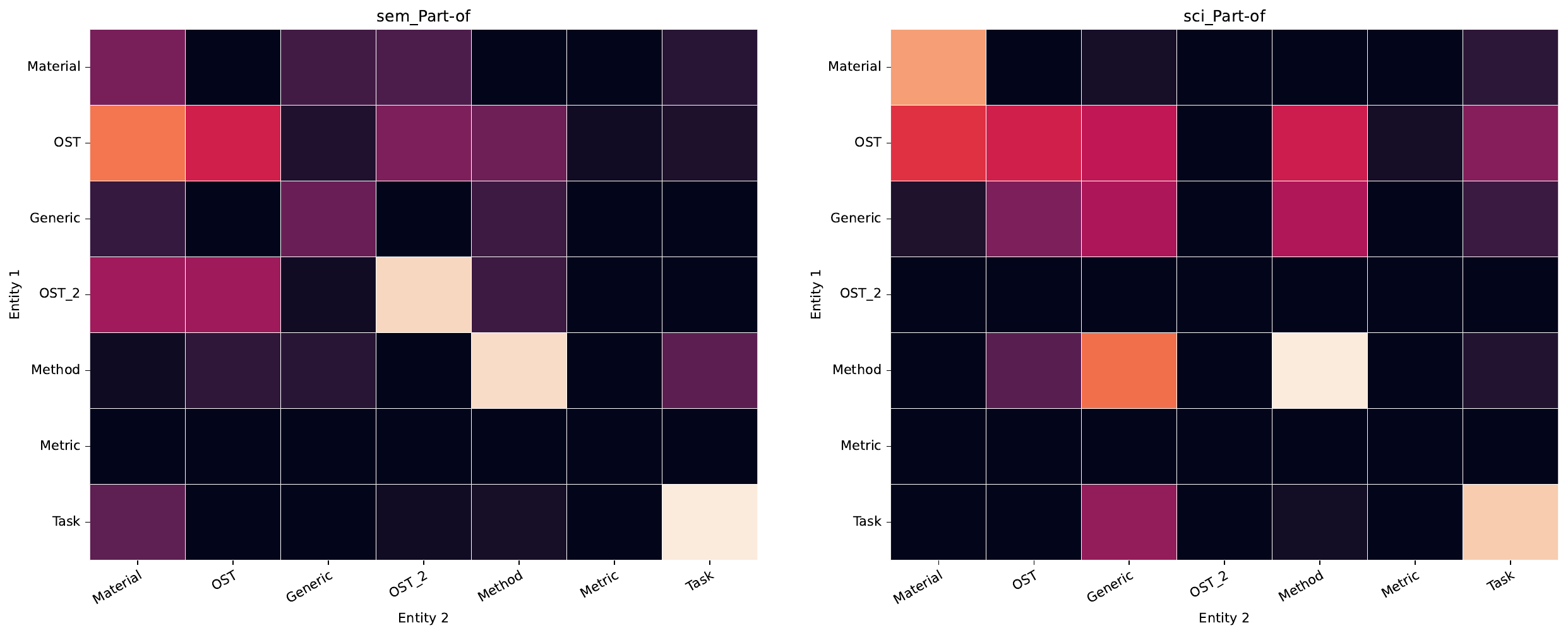}
\caption{The co-occurrence distribution of ``Part-of/Part-whole''  in two overlapped corpora.}
\label{occurrence-Part-of}
\end{figure*}

\end{document}